# A Relevant Reduction Method for Dynamic Modeling of a Seven-linked Humanoid Robot in the Three-dimensional Space

Amira Aloulou[a], Olfa Boubaker[a],*

[a] *National Institute of Applied Sciences and Technology, Centre Urbain Nord, BP 676 - 1080 Tunis, Tunisia*

**Abstract**

In this paper, the dynamic modeling of a seven linked humanoid robot, is accurately developed, in the three dimensional space using the Newton-Euler formalism. The aim of this study is to provide a clear and a systematic approach so that starting from generalized motion equations of all rigid bodies of the humanoid robot one can establish a reduced dynamical model. The resulting model can be expended either for simulation propositions or implemented for any given control law. In addition, transformations and developments, proposed here, can be exploited for modeling any other three-dimensional humanoid robot with a different morphology and variable number of rigid bodies and degrees of freedom.

## 1. Introduction

The study and materialization of humanoid robots is a wild field of investigation where dynamic modeling plays a crucial role since all following stages of implementation are narrowly depending on it and especially the synthesis of an efficient control law. Indeed, the dynamic model has to be accurate by taking account of all forces and torques involved in the dynamic motion of the robot. Determining the degrees of freedom's number and location and getting the kinematic model are the two prerequisites needed to establish a dynamic modeling. At this point, the Lagrange and Newton–Euler approaches represent the main methodologies employed allowing the dynamic modeling of any given robotic system [1]. The Lagrangian formalism gives a simple representation of a system having several joints by supplying at every instant generalized coordinates of all particles involved in the robotic system motion [2]. However, this method shows certain limitations. For complex systems, choosing coordinates for the particles during motion is a difficult task. Moreover, this approach doesn't provide direct data regarding the required torques needed to activate each link. Especially, the implementation of some complex links involving a friction as in the double-support-phase (DSP) walking cycle becomes quite unrealizable in a purely Lagrangian context. Finally, the estimation and inversion of certain Jacobians may produce nonlinear and dense inertia operators thus generating a high calculation cost [3]. Therefore, the Newton-Euler methodology has been chosen to lead our study. It is based on the observation and determination of velocities in one or more given observation points. These velocities represent particles velocities when moving through the observation point at different times [4]. Thus, observation points stand for current positions of particles at a given instant. Even if the Newton-Euler formalism involves taking account of more parameters than the Lagrangian one in the case of systems composed of many rigid bodies, the determination and inversion of the inertia operators doesn't imply a high calculation cost.

---

* Corresponding author. Tel.:+0-216-1-703-929; fax: +0-216-1-704-929.
*E-mail address:* olfa.boubaker@insat.rnu.tn.

With regard to other works dealing with the dynamic modeling of bipedal robots in the three dimensional space, we may notice the lack of papers introducing a clear and sequential methodology to follow in order to establish a reduced dynamic model so that it can be directly expended for simulations and implemented for any given control law [5]. Throughout this article, our aim is then to propose a plain and exhaustive method based on the Newton-Euler formalism to obtain a satisfactory three-dimensional dynamic modeling whatever number of rigid bodies or degrees of freedom is considered. Nevertheless, the lower body morphology of a female humanoid robot prototype introduced in a previous paper [6] will be adopted to realize this demonstration. This robot prototype is composed of seven links associated to 12 DOF. In previous works [6], all physical parameters corresponding to each link have been determined and a kinematic model has been established. In this paper, generalized three dimensional motion equations for the rotation and the translation of each link are reached. A first dynamic modeling corresponding to the free robotic system when achieving the single support phase is proposed via some rigorously explained transformations and developments. We show how to finally get reduced free or constraint three dimensional dynamic models. Both free and constraint resulting dynamic models are directly expendable for reaching a whole walking cycle including the single support, impact and double support phases.

## 2. Dynamic Modeling

The bipedal robot prototype [7] is composed of seven links associated to 12 DOF. Each rigid body Ci of the humanoid robot is characterized by its own physical parameters determined using Winter statistical model and detailed in [6]. As a prerequisite to dynamic modeling in the three dimensional space, the three-dimensional kinematic model basically founded on Euler's transformation matrix has been established and then the dynamic model will be computed using the Newton-Euler method.

### 2.1. Newton-Euler Principle for Dynamic Modeling

For each link Ci, generalized motion equations for the rotation and the translation are used such as [7]:

$$I_i \dot{w}_i = f_i + F_i + F_{i+1} + G_i + G_{i+1} + \tau_i + \tau_{i+1} \tag{1}$$

$$M_i \ddot{X}_i = M_i g + \Gamma_i - \Gamma_{i+1} \tag{2}$$

where each term of (1) and (2) is described in the nomenclature given below.

**Nomenclature**

$W_i, \dot{W}_i$    Angular position and velocity of the link Ci

$X_i, \dot{X}_i, \ddot{X}_i$    Linear position, velocity and acceleration of the link Ci

$F_i, F_{i+1}$    Torques due to the holonom force applied respectively to the proximal and distal articulation of the link Ci expressed in the body coordinate system

$G_i, G_{i+1}$    Non-holonom torques applied respectively to the proximal and distal articulation of the link Ci expressed in the body coordinate system

$\tau_i, \tau_{i+1}$    Muscular torques applied respectively to the proximal and distal articulation of the link Ci expressed in the body coordinate system

$\Gamma_i, \Gamma_{i+1}$    Holonom forces applied respectively to the proximal and distal articulation of the link Ci expressed in the inertial coordinate system

$f_i$    Intrinsic torque of the link Ci expressed in the body coordinates system $(x_i, y_i, z_i)$ and relating angular velocity to the link inertia

$\Lambda_i$    Non-holonom torque applied to the proximal articulation of the link Ci expressed in the body coordinate system of the previous link Ci-1

$\Lambda_{i+1}$    Non-holonom torque applied to the proximal articulation of the link Ci expressed in the body coordinate system of the previous link Ci-1

$R_i, R_{i+1}$    Non holonom torque applied to the distal articulation of the link Ci expressed in the body coordinate system

$T_i$    Torque applied to the proximal articulation of the link Ci expressed in the body coordinate system of the previous link Ci-1

$T_{i+1}$    Torque applied to the distal articulation of the link Ci expressed in the body coordinate system

Human body's balance of forces and torques reveals that humanoid limbs are subject to three kinds of forces: holonom, non holonom and mechanic forces. Holonom forces and torques result of the interaction between limbs whereas non-holonom torques consist in the effort a limb is subject to in order to remain aligned with the previous limb. Finally, mechanical torques are muscular torques applied by human body's muscles to move limbs. Fig.1 presents the lower body joints and links whereas Fig.2 shows the applied forces and torques to the considered humanoid robot lower body.

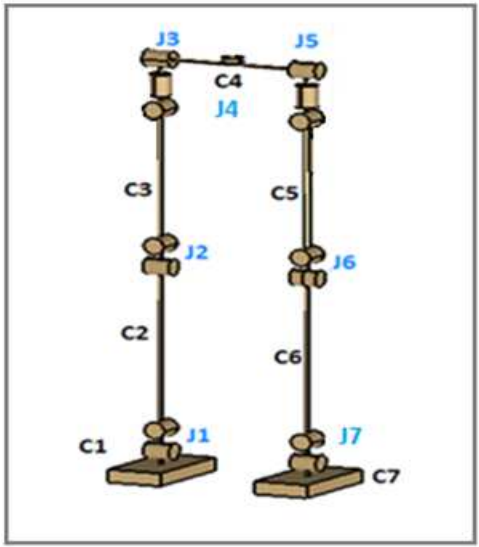

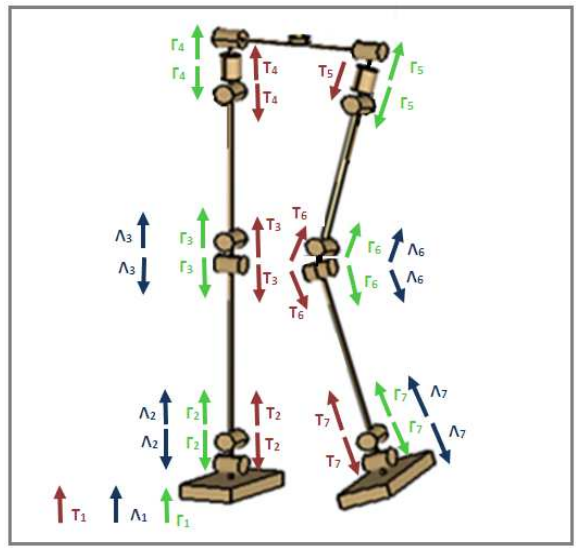

Fig.1: Links and Joints of the Bipedal Robot

Fig.2: Applied forces and torques to the Bipedal Robot

*2.2. Constraints*

A humanoid robot is always subject to two kinds of constraints: non holonomic constraints and holonomic ones. For the case study of the seven-linked robot, we get the following constraints:

**Non holonomic constraints:** We have enumerated four non holonomic constraints that are to be found at the knees and the ankles. The first constraint occurs when the bipedal robot is standing on its right foot. In that posture, the right leg can only rotate in the sagittal plane at the right ankle level. Therefore, there is only one degree of freedom at the right ankle and two degrees of connection. In the same way and for the second constraint, there is a similar configuration at the level of the left ankle. For each rigid body subject to a non holonomic constraint, the Ri matrix represents the degrees of connection involved while the Qi matrix contains the degrees of freedom. Regarding these two constraints, Ri and Qi are the following:

$$R_2 = R_7 = \begin{bmatrix} 1 & 0 & 0 \\ 0 & 0 & 1 \end{bmatrix}^T \quad \text{and} \quad Q_2 = Q_7 = \begin{bmatrix} 0 & 1 & 0 \end{bmatrix}^T$$

The third constraint is to be found in the region of the right knee. In that location, there is only one degree of freedom and two degrees of connection. Similarly for the fourth constraint, there is an identical configuration at the level of the left knee. Ri and Qi for these two last constraints are given by:

$$R_3 = R_6 = \begin{bmatrix} 1 & 0 & 0 \\ 0 & 0 & 1 \end{bmatrix}^T \quad \text{and} \quad Q_3 = Q_6 = \begin{bmatrix} 0 & 1 & 0 \end{bmatrix}^T$$

When translating the previous non holonomic constraints into equations, we get the following system:

$$\begin{cases} R_2^T(A_1^T A_2 W_2 - W_1) = 0 \\ R_3^T(A_2^T A_3 W_3 - W_2) = 0 \\ R_6^T(A_5^T A_6 W_6 - W_5) = 0 \\ R_7^T(A_6^T A_7 W_7 - W_6) = 0 \end{cases} \quad (3)$$

**Holonomic constraints:** Twenty-one holonomic constraints directly arise from the bipedal robot kinematic modeling given in [6]. They may be described as constraints of connection between segments in order to maintain a rigid body linked to both the previous and next link. These constraints are enumerated in the system of equations given by (4):

$$\begin{cases} X_1 = A_1 K_1 \\ X_2 = A_2 K_2 - A_1 L_1 + X_1 \\ X_3 = A_3 K_3 - A_2 L_2 + X_2 \\ X_4 = A_4 K_4 - A_3 L_3 + X_3 \\ X_5 = A_5 K_5 - A_4 L_4 + X_4 \\ X_6 = A_6 K_6 - A_5 L_5 + X_5 \\ X_7 = A_7 K_7 - A_6 L_6 + X_6 \end{cases} \quad (4)$$

*2.3. A method for Dynamic Modeling*

Holonom forces and muscular torques are applied to every articulation whereas non holonom forces are only to be found in the articulations of ankles $J_1$ and $J_7$ and knees $J_2$ and $J_6$. Table 1 (resp. 2) is a summary of the forces and torques applied to the seven rigid bodies composing the bipedal robot for the rotation (resp. translation):

Table 1: Forces and torques applied to the bipedal robot for the equation of rotation

| Body i | $f_i$ | $F_i$ | $F_{i+1}$ | $G_i$ | $G_{i+1}$ | $\tau_i$ | $\tau_{i+1}$ |
|---|---|---|---|---|---|---|---|
| 1 | f1 | $K_1 x A_1^T \Gamma_1$ | $-L_1 x A_1^T \Gamma_2$ | 0 | $-R_2 \Lambda_2$ | $A_1^T T_1$ | $-T_2$ |
| 2 | f2 | $K_2 x A_2^T \Gamma_2$ | $-L_2 x A_2^T \Gamma_3$ | $A_2^T A_1 R_2 \Lambda_2$ | $-R_3 \Lambda_3$ | $A_2^T A_1 T_2$ | $-T_3$ |
| 3 | f3 | $K_3 x A_3^T \Gamma_3$ | $-L_3 x A_3^T \Gamma_4$ | $A_3^T A_2 R_3 \Lambda_3$ | 0 | $A_3^T A_2 T_3$ | $-T_4$ |
| 4 | f4 | $K_4 x A_4^T \Gamma_4$ | $L_4 x A_4^T \Gamma_5$ | 0 | 0 | $A_4^T A_3 T_4$ | $-T_5$ |
| 5 | f5 | $-K_5 x A_5^T \Gamma_5$ | $L_5 x A_5^T \Gamma_6$ | 0 | $R_6 \Lambda_6$ | $-A_5^T A_4 T_5$ | $T_6$ |
| 6 | f6 | $-K_6 x A_6^T \Gamma_6$ | $L_6 x A_6^T \Gamma_7$ | $-A_6^T A_5 R_6 \Lambda_6$ | $R_7 \Lambda_7$ | $-A_6^T A_5 T_6$ | $T_7$ |
| 7 | f7 | $-K_7 x A_7^T \Gamma_7$ | 0 | $-A_7^T A_6 R_7 \Lambda_7$ | 0 | $-A_7^T A_6 T_7$ | 0 |

Table 2: Forces and torques applied to the bipedal robot for the equation of translation

| Body i | $M_i.g$ | $\Gamma_i$ | $\Gamma_{i+1}$ |
|---|---|---|---|
| 1 | $M_1 g$ | $\Gamma_1$ | $-\Gamma_2$ |
| 2 | $M_2 g$ | $\Gamma_2$ | $-\Gamma_3$ |
| 3 | $M_3 g$ | $\Gamma_3$ | $-\Gamma_4$ |
| 4 | $M_4 g$ | $\Gamma_4$ | $\Gamma_5$ |
| 5 | $M_5 g$ | $-\Gamma_5$ | $\Gamma_6$ |
| 6 | $M_6 g$ | $-\Gamma_6$ | $\Gamma_7$ |
| 7 | $M_7 g$ | $-\Gamma_7$ | 0 |

Based on generalized motion equations (1) and (2), results provided by Tables 1 and 2 can be written after simplification as follows:

$$\begin{cases} I_1 \dot{W}_1 = f_1 + KK_1 A_1^T \Gamma_1 - LL_1 A_1^T \Gamma_2 - R_2 \Lambda_2 + A_1^T T_1 - T_2 \\ I_2 \dot{W}_2 = f_2 + KK_2 A_2^T \Gamma_2 - LL_2 A_2^T \Gamma_3 - R_3 \Lambda_3 + A_2^T A_1 R_2 \Lambda_2 + A_2^T A_1 T_2 - T_3 \\ I_3 \dot{W}_3 = f_3 + KK_3 A_3^T \Gamma_3 - LL_3 A_3^T \Gamma_4 + A_3^T A_2 R_3 \Lambda_3 + A_3^T A_2 T_3 - T_4 \\ I_4 \dot{W}_4 = f_4 + KK_4 A_4^T \Gamma_4 + LL_4 A_4^T \Gamma_5 + A_4^T A_3 T_4 - T_5 \\ I_5 \dot{W}_5 = f_5 - KK_5 A_5^T \Gamma_5 + LL_5 A_5^T \Gamma_6 + R_6 \Lambda_6 - A_5^T A_4 T_5 + T_6 \\ I_6 \dot{W}_6 = f_6 - KK_6 A_6^T \Gamma_6 + LL_6 A_6^T \Gamma_7 + R_7 \Lambda_7 - A_6^T A_5 R_6 \Lambda_6 - A_6^T A_5 T_6 + T_7 \\ I_7 \dot{W}_7 = f_7 - KK_7 A_7^T \Gamma_7 - A_7^T A_6 R_7 \Lambda_7 - A_7^T A_6 T_7 + T_7 \end{cases} \quad (5.a)$$

$$\begin{cases} M_1\ddot{X}_1 = M_1 g + \Gamma_1 - \Gamma_2 \\ M_2\ddot{X}_2 = M_2 g + \Gamma_2 - \Gamma_3 \\ M_3\ddot{X}_3 = M_3 g + \Gamma_3 - \Gamma_4 \\ M_4\ddot{X}_4 = M_4 g + \Gamma_4 + \Gamma_5 \\ M_5\ddot{X}_5 = M_5 g - \Gamma_5 + \Gamma_6 \\ M_6\ddot{X}_6 = M_6 g - \Gamma_6 + \Gamma_7 \\ M_7\ddot{X}_7 = M_7 g - \Gamma_7 \end{cases} \qquad (5.b)$$

In Table 1, it may be noticed that most involved forces include in their expressions a vector product. By using antisymmetric matrices to eliminate vector products, every equation for rotation is reduced as shown in (5). Indeed, the role of an antisymmetric matrix [8] is to conceal the vector products by replacing them with simpler products. Assuming that $z_1 \in \Re^{42 \times 1}$ is the state vector of the system composed of angular and linear accelerations and $\Gamma \in \Re^{21 \times 1}, \Lambda \in \Re^{8 \times 1}$ and $T \in \Re^{21 \times 1}$ are the holonom, non-holonom and muscular forces and torques respectively, the system (5) may be written as:

$$P_1 z_1 = P_2 + P_3 \Gamma + P_4 \Lambda + P_5 T \qquad (6)$$

where $P_1$, $P_2$, $P_3$, $P_4$ and $P_5$ are matrices of appropriate dimensions including, on the one hand, the different products between proximal/distal distances and Euler's transformation matrices and on the other hand inertias and masses of the rigid body whose forces and torques are at play. To exploit the results of non holonomic constraints, we will differentiate (3) with respect to time which gives:

$$\begin{cases} R_2^T(-\dot{W}_1 - WW_1 A_1^T A_2 W_2 + A_1^T A_2 \dot{W}_2) = 0 \\ R_3^T(-\dot{W}_2 - WW_2 A_2^T A_3 W_3 + A_2^T A_3 \dot{W}_3) = 0 \\ R_6^T(-\dot{W}_5 - WW_5 A_5^T A_6 W_6 + A_5^T A_6 \dot{W}_6) = 0 \\ R_7^T(-\dot{W}_6 - WW_6 A_6^T A_7 W_7 + A_6^T A_7 \dot{W}_7) = 0 \end{cases}$$

When arranging the above system of equations into a more compact form, the following equation is obtained:

$$P_4^T z_1 = P_6 \qquad \text{with} \quad P_6 = \begin{pmatrix} R_2^T WW_1 A_1^T A_2 W_2 \\ R_3^T WW_2 A_2^T A_3 W_3 \\ R_6^T WW_5 A_5^T A_6 W_6 \\ R_7^T WW_6 A_6^T A_7 W_7 \end{pmatrix}, \quad P_6 \in \Re^{8 \times 1} \qquad (7)$$

Let's come-back to the system (4) that summarizes all holonomic constraints applied to the humanoid biped and study the case of the first rigid body C1. When differentiating the Euler's transformation matrix $A_1$ with respect to time, the resulting equation is (8):

$$\dot{A}_1 = \frac{dA_1}{dt} = A_1 WW_1 \qquad (8)$$

Using the result given by (8) and differentiating the first equation of (5) with respect to time, the result is:

$$\dot{X}_1 - A_1 WW_1 K_1 = 0$$

If the first equation of (5) is differentiated with respect to time twice, the following equation is obtained:

$$\ddot{X}_1 - \frac{d(A_1 WW_1 K_1)}{dt} = \ddot{X}_1 - A_1 (WW_1)^2 K_1 + A_1 KK_1 \dot{W}_1 = 0$$

In order to apply the same reasoning to the six other rigid bodies, one has to calculate time derivatives of Euler's transformation matrices $A_2, A_3, A_4, A_5, A_6$ and $A_7$ and then implement results into time derivatives of system (5) so that:

$$\begin{cases} \ddot{X}_1 + A_1 KK_1 \dot{W}_1 = A_1 (WW_1)^2 K_1 \\ \ddot{X}_2 + A_2 KK_2 \dot{W}_2 - A_1 LL_1 \dot{W}_1 - \ddot{X}_1 = A_2 (WW_2)^2 K_2 - A_1 (WW_1)^2 L_1 \\ \ddot{X}_3 + A_3 KK_3 \dot{W}_3 - A_2 LL_2 \dot{W}_2 - \ddot{X}_2 = A_3 (WW_3)^2 K_3 - A_2 (WW_2)^2 L_2 \\ \ddot{X}_4 + A_4 KK_4 \dot{W}_4 - A_3 LL_3 \dot{W}_3 - \ddot{X}_3 = A_4 (WW_4)^2 K_4 - A_3 (WW_3)^2 L_3 \\ -\ddot{X}_5 - A_5 KK_5 \dot{W}_5 + A_4 LL_4 \dot{W}_4 + \ddot{X}_4 = -A_5 (WW_5)^2 K_5 + A_4 (WW_4)^2 L_4 \\ -\ddot{X}_6 - A_6 KK_6 \dot{W}_6 + A_5 LL_5 \dot{W}_5 + \ddot{X}_5 = -A_6 (WW_6)^2 K_6 + A_5 (WW_5)^2 L_5 \\ -\ddot{X}_7 - A_7 KK_7 \dot{W}_7 + A_6 LL_6 \dot{W}_6 + \ddot{X}_6 = -A_7 (WW_7)^2 K_7 + A_6 (WW_6)^2 L_6 \end{cases} \quad (9)$$

Equation (10) is obtained by arranging the system (9) into a more compact form:

$$P_3^T z_1 = P_7 \quad \text{with } P_7 = \begin{pmatrix} A_1 (WW_1)^2 K_1 \\ A_2 (WW_2)^2 K_2 - A_1 (WW_1)^2 L_1 \\ A_3 (WW_3)^2 K_3 - A_2 (WW_2)^2 L_2 \\ A_4 (WW_4)^2 K_4 - A_3 (WW_3)^2 L_3 \\ -A_5 (WW_5)^2 K_5 + A_4 (WW_4)^2 L_4 \\ -A_6 (WW_6)^2 K_6 + A_5 (WW_5)^2 L_5 \\ -A_7 (WW_7)^2 K_7 + A_6 (WW_6)^2 L_4 \end{pmatrix}, \quad P_7 \in \Re^{21 \times 1} \quad (10)$$

When combining equations (7) and (10), the following equality is reached:

$$[P_3 \ P_4]^T z_1 = [P_7^T \ P_6^T]^T \quad (11)$$

$P_1$ is an inversible matrix, $z_1$ can then be expressed according to other terms of the equality (6):

$$z_1 = P_1^{-1} [P_2 + P_3 \Gamma + P_4 \Lambda + P_5 T] \quad (12)$$

Replacing $z_1$ in (11) by its expression in (12) and then isolating it in the left side of the equation, (13) is reached:

$$\begin{bmatrix} \Gamma \\ \Lambda \end{bmatrix} = \left\{ [P_3 \ P_4]^T P_1^{-1} [P_3 \ P_4] \right\}^{-1} [P_7^T \ P_6^T]^T - [P_3 \ P_4]^{-1} (P_2 + P_5 T) \quad (13)$$

In (13), holonomic force $\Gamma$ and non holonomic torque $\Lambda$ are expressed as state variables. Knowing that $\Gamma \in \Re^{21 \times 1}$ and $\Lambda \in \Re^{8 \times 1}$, the formulation (13) presents a system composed of 29 state variables instead of the initial 42 ones. This already represents a reduction of the DOF involved. However, some transformations are still needed to be proceeded to get a system with a minimal number of state variables. Assuming that $P_3^T = [P_8 \ P_9]$, the following equality is obtained:

$$P_3^T z_1 = [P_8 \ P_9] z_1 = P_8 \dot{W} + P_9 \ddot{X}$$

If $P_3$ is replaced by its expression in equation (10), the following equation is reached:

$$P_8 \dot{W} + P_9 \ddot{X} = P_7 \quad (14)$$

To reach a state modeling from previous equation (11), vector $\ddot{X}$ is isolated and expressed according to the remaining elements of the equation so that:

$$\ddot{X} = -P_9^{-1} P_8 \dot{W} + P_9^{-1} P_7$$

which can be written as:

$$\begin{pmatrix} \dot{W} \\ \ddot{X} \end{pmatrix} = \begin{pmatrix} I_{21} \\ -P_9^{-1} P_8 \end{pmatrix} \dot{W} + \begin{pmatrix} 0_{21} \\ P_9^{-1} P_7 \end{pmatrix} \quad \text{where } o_{21} \in \Re^{21 \times 1} \text{ is the null vector and } I_{21} \in \Re^{21 \times 21} \text{ is the identity matrix.}$$

Let's assume that:

$$P_{10} = \begin{pmatrix} I_{21} \\ -P_9^{-1} P_8 \end{pmatrix}, \quad P_{10} \in \Re^{42 \times 21} \qquad \text{and} \qquad P_{11} = \begin{pmatrix} 0_{21} \\ P_9^{-1} P_7 \end{pmatrix}, \quad P_{11} \in \Re^{42 \times 1}$$

It can be therefore written that:

$$z_1 = P_{10}\dot{W} + P_{11} \tag{15}$$

Inserting the expression of $z_1$ given in (15) into the equation of system (6) and then isolating the first left side term in the resulting equation, one gets:

$$P_1 P_{10} \dot{W} = -P_1 P_{11} + P_2 + P_3 \Gamma + P_4 \Lambda + P_5 T$$

Let's multiply the previous equality by $P_{10}{}^T$ so that:

$$P_{10}{}^T P_1 P_{10} \dot{W} = -P_{10}{}^T P_1 P_{11} + P_{10}{}^T P_2 + P_{10}{}^T P_3 \Gamma + P_{10}{}^T P_4 \Lambda \tag{16}$$

Verifying that $P_{10}^T P_3 \Gamma = 0$, equation (16) becomes:

$$P_{10}{}^T P_1 P_{10} \dot{W} = -P_{10}{}^T P_1 P_{11} + P_{10}{}^T P_2 + P_{10}{}^T P_4 \Lambda + P_{10}{}^T P_5 T \tag{17}$$

Angular accelerations $\dot{W}_1$ to $\dot{W}_7$ defined in [6] may be written under a matricial form as:

$$\dot{W} = P_{12} \ddot{\Phi} + P_{13} \tag{18}$$

Where $\ddot{\Phi}$ is the second time derivative of the vector $\Phi$ that contains positions of only active DOFs. Then, using expression (18) in equation (17), the following equality is obtained:

$$P_{10}{}^T P_1 P_{10}(P_{12} \ddot{\Phi} + P_{13}) = -P_{10}{}^T P_1 P_{11} + P_{10}{}^T P_2 + P_{10}{}^T P_4 \Lambda + P_{10}{}^T P_5 T \tag{19}$$

Let's isolate the first left side term of the previous equality and multiply the whole equality by $P_{12}{}^T$:

$$P_{12}{}^T P_{10}{}^T P_1 P_{10} P_{12} \ddot{\Phi} = -P_{12}{}^T P_{10}{}^T P_1 P_{11} + P_{12}{}^T P_{10}{}^T P_2 + P_{12}{}^T P_{10}{}^T P_4 \Lambda + P_{12}{}^T P_{10}{}^T P_5 T - P_{12}{}^T P_{10}{}^T P_1 P_{10} P_{13}$$

It is finally shown that $P_{12}^T P_{10}^T P_4 \Lambda = 0$ consequently the below equation may be written as follows:

$$P_{12}{}^T P_{10}{}^T P_1 P_{10} P_{12} \ddot{\Phi} = -P_{12}{}^T P_{10}{}^T P_1 P_{11} + P_{12}{}^T P_{10}{}^T P_2 + P_{12}{}^T P_{10}{}^T P_5 T - P_{12}{}^T P_{10}{}^T P_1 P_{10} P_{13} \tag{20}$$

In order to get closer to the motion equation general form, all terms depending on the state variable $\Phi$ and its first derivative $\dot{\Phi}$ and second derivative $\ddot{\Phi}$ are placed on one side of the equality and all remaining terms on the other side. Equation (20) becomes then:

$$P_{12}{}^T P_{10}{}^T P_1 P_{10} P_{12} \ddot{\Phi} + P_{12}{}^T P_{10}{}^T P_1 (P_{11} + P_{10} P_{13}) - P_{12}{}^T P_{10}{}^T P_2 = P_{12}{}^T P_{10}{}^T P_5 T \tag{21}$$

The final reached equation (21) stands for the dynamic equation of a free robotic system. This last expression is quite conformed to the general form of a robotic dynamic modeling given in (22):

$$J(\Phi)\ddot{\Phi} + H(\Phi, \dot{\Phi}) + G(\Phi) = D.\tau \tag{22}$$

where $\Phi$ is the state vector describing the dynamics of the humanoid robot system, $J(\Phi)$ is the positive definite inertia matrix, $H(\Phi, \dot{\Phi})$ is the vector of the Coriolis and centripetal torques, $G(\Phi)$ is the positive definite gravity vector, D is a nonsingular input map matrix and $\tau$ is the vector of control inputs. Making correspondence between equations (21) and (22), the following system is obtained:

$$\begin{cases} J(\Phi) = P_{12}^T P_{10}^T P_1 P_{10} P_{12}, & J(\Phi) \in \Re^{12*12} \\ H(\Phi, \dot{\Phi}) = P_{12}^T P_{10}^T (P_1 P_{11} + P_1 P_{10} P_{13}), & H(\Phi, \dot{\Phi}) \in \Re^{12*1} \\ G(\Phi) = -P_{12}^T P_{10}^T P_2, & G(\Phi) \in \Re^{12*1} \\ D = P_{12}^T P_{10}^T P_5, & D \in \Re^{12*12} \end{cases} \tag{23}$$

*2.4. Resulting Constraint Dynamic Model*

Whether considering the impact phase or the double support phase, the general robotic constraint dynamic model of the bipedal robot is described by [5]:

$$J(\Phi)\ddot{\Phi} + H(\Phi, \dot{\Phi}) + G(\Phi) = D\tau + \frac{\partial E^T}{\partial \Phi} \Gamma_c \qquad (24)$$

where E is the contact point defined such as $E = E(\Phi) = [E_x \quad E_y \quad E_z]^T$ and $\Gamma_c$ is the contact force with the ground. During the ground contact, the free end of the biped, at the completion of the step, comes into contact with the ground. This phase is assumed to take place in an infinitesimal time interval. The impact force is described by:

$$\Gamma_{c\ imp} = -\frac{\partial E}{\partial \Phi}\Big|_{\Phi_{imp}} J(\Phi_{imp})\dot{\Phi}(t_0^-) \qquad (25)$$

where $t_0^-$ is the instant that directly occurs before the impact. During the double support phase the reaction force is described by:

$$\Gamma_{c\ cont} = \left[\left(\frac{\partial E}{\partial \Phi}J^{-1}(\Phi)\frac{\partial E^T}{\partial \Phi}\right)\left(\frac{\partial E}{\partial \Phi}J^{-1}(\Phi)\left[H(\Phi,\dot{\Phi}) + G(\Phi) - D\tau\right] - \frac{\partial}{\partial \Phi}\left(\frac{\partial E}{\partial \Phi}\dot{\Phi}\right)\dot{\Phi}\right)\right]^{-1} \qquad (26)$$

Various solutions may be used to solve the position/ force control problem [9, 10, 11]**.** For our case two control laws have been already applied to the reduced model: a feedback linearizing control [7] and a Minimum Jerk-based one [12]. Satisfactory simulation results have been obtained for the both cases.

## 3. Conclusion

In this paper, the three dimensional dynamic modeling of a seven linked humanoid robot has been accurately developed using the Newton-Euler formalism. A clear and sequential methodology is provided to establish a reduced and expendable model. The reduced model allows a direct use for control law implementation.